\pdfoutput=1
\PassOptionsToPackage{utf8}{inputenc}
\documentclass{bioinfo}

\usepackage{pifont}
\usepackage{booktabs}       
\usepackage{url}            
\usepackage{amsfonts}       
\usepackage{nicefrac}       
\usepackage{amsmath}
\usepackage{graphicx}
\usepackage{bm, upgreek}
\usepackage{amssymb}
\usepackage{array}
\usepackage{multirow}
\usepackage{mathtools}
\usepackage{arydshln}
\usepackage{tabularx}
\usepackage{color}
\usepackage{soul}
\usepackage{caption}
\usepackage{xcolor}
\usepackage{comment}
\usepackage{adjustbox}
\usepackage{hyperref}
\usepackage{cleveref}
\usepackage[ruled,vlined]{algorithm2e}
\usepackage{makecell}
\usepackage{diagbox}
\usepackage{subfig}
\usepackage{xcolor, soul}

\newcommand{\ours}{\textsc{ConNER}}

\definecolor{readablered}{RGB}{170,0,0}

\copyrightyear{2022} \pubyear{2022}

\access{Advance Access Publication Date: Day Month Year}
\appnotes{Application Note}
\begin{document}
\firstpage{1}

\subtitle{Data and text mining}

\title[ConNER]{Enhancing Label Consistency on Document-level Named Entity Recognition}
\author[Jeong \textit{et~al}.]{Minbyul Jeong\,$^{\text{\sfb 1}}$ and Jaewoo Kang\,$^{\text{\sfb 1,2,3,*}}$}
\address{$^{\text{\sf 1}}$Department of Computer Science and Engineering, Korea University, Seoul, 02841, Republic of Korea \\
$^{\text{\sf 2}}$Interdisciplinary Graduate Program in Bioinformatics, Korea University, Seoul, Republic of Korea \\
$^{\text{\sf 3}}$AIGEN Sciences, Seoul, 04778, Republic of Korea}

\corresp{
$^\ast$To whom correspondence should be addressed.
}

\history{Received on XXXXX; revised on XXXXX; accepted on XXXXX}

\editor{Associate Editor: XXXXXXX}

\abstract{
\textbf{Summary:}
Named entity recognition (NER) is a fundamental part of extracting information from documents in biomedical applications.
A notable advantage of NER is its consistency in extracting biomedical entities in a document context.
Although existing document NER models show consistent predictions, they still do not meet our expectations.
We investigated whether the adjectives and prepositions within an entity cause a low label consistency, which results in inconsistent predictions.
In this paper, we present our method, \textit{ConNER}, which enhances the label dependency of modifiers (e.g., adjectives and prepositions) to achieve higher label agreement. 
ConNER refines the draft labels of the modifiers to improve the output representations of biomedical entities.
The effectiveness of our method is demonstrated on four popular biomedical NER datasets; in particular, its efficacy is proved on two datasets with 7.5–8.6\% absolute improvements in the F1 score.
We interpret that our ConNER method is effective on datasets that have intrinsically low label consistency.
In the qualitative analysis, we demonstrate how our approach makes the NER model generate consistent predictions.
\\
\textbf{Availability and implementation:}
Our code and resources are available at \href{https://github.com/dmis-lab/ConNER/}{https://github.com/dmis-lab/ConNER/}.\\
\textbf{Contact:}
\href{kangj@korea.ac.kr}{kangj@korea.ac.kr} \\
\textbf{Supplementary information:} Supplementary data are available at \textit{Bioinformatics}
online.}

\maketitle
\section{Introduction}
Named entity recognition (NER) is the task of determining entity boundaries and classifying categories of named entities.
NER is a fundamental part of biomedical applications~\citep{kim2019neural, wei2019pubtator, lee2020answering, weber2021hunflair, lewis2021paq, sung2022bern2}.
In the general domain, recent studies have attempted to train and evaluate NER models in a document-level context~\citep{yu2020named, yamada2020luke, wang2021improving, gui2021leveraging}.
Likewise, the biomedical domain has shifted its focus on evaluating document contexts rather than sentence contexts~\citep{wei2019pubtator, wang2021improving, weber2021hunflair}.

There are several advantages of using document NER models: (1) The models suggest a better way to bridge the gap between research and application fields.
Following previous studies, several researches have leveraged sentence NER models in biomedical domains~\citep{cho2019biomedical, perera2020named, jeong2021regularization}.
However, biomedical applications require an evaluation of the document rather than the sentence context~\citep{wei2019pubtator, kim2019neural, weber2021hunflair, sung2022bern2}.
(2) Document NER models provide proper and consistent predictions owing to context completeness.
Recent works~\citep{yamada2020luke, wang2021improving,  gui2021leveraging} have shown that using document contexts improves the accuracy of the models.
Although the authors of~\citep{gui2021leveraging} used powerful context representations such as BERT~\citep{devlin2018bert} or ELMo~\citep{peters-etal-2018-deep}, their inability to model document-level label consistency resulted in insufficient performance.
Therefore, it is challenging to understand which factors contribute to creating \textit{document} NER models that produce a \textit{consistent} prediction in the same manner.

To tackle the above challenges, a series of studies \citep{fu2020interpretable, fu2021larger} have provided an interpretable evaluation to identify the attributes of datasets depending on their characteristics.
Our intuitive motivation is that entity-aware attributes are beneficial for achieving higher label consistency, which can also improve the accuracy of the NER models.
In this paper, our goal is to develop a NER model that can be trained to predict an entity consistently in a document context. 

\begin{figure*}[t]
  \centering
  \vspace{-1.25cm}
  \includegraphics[width=0.63\textwidth]{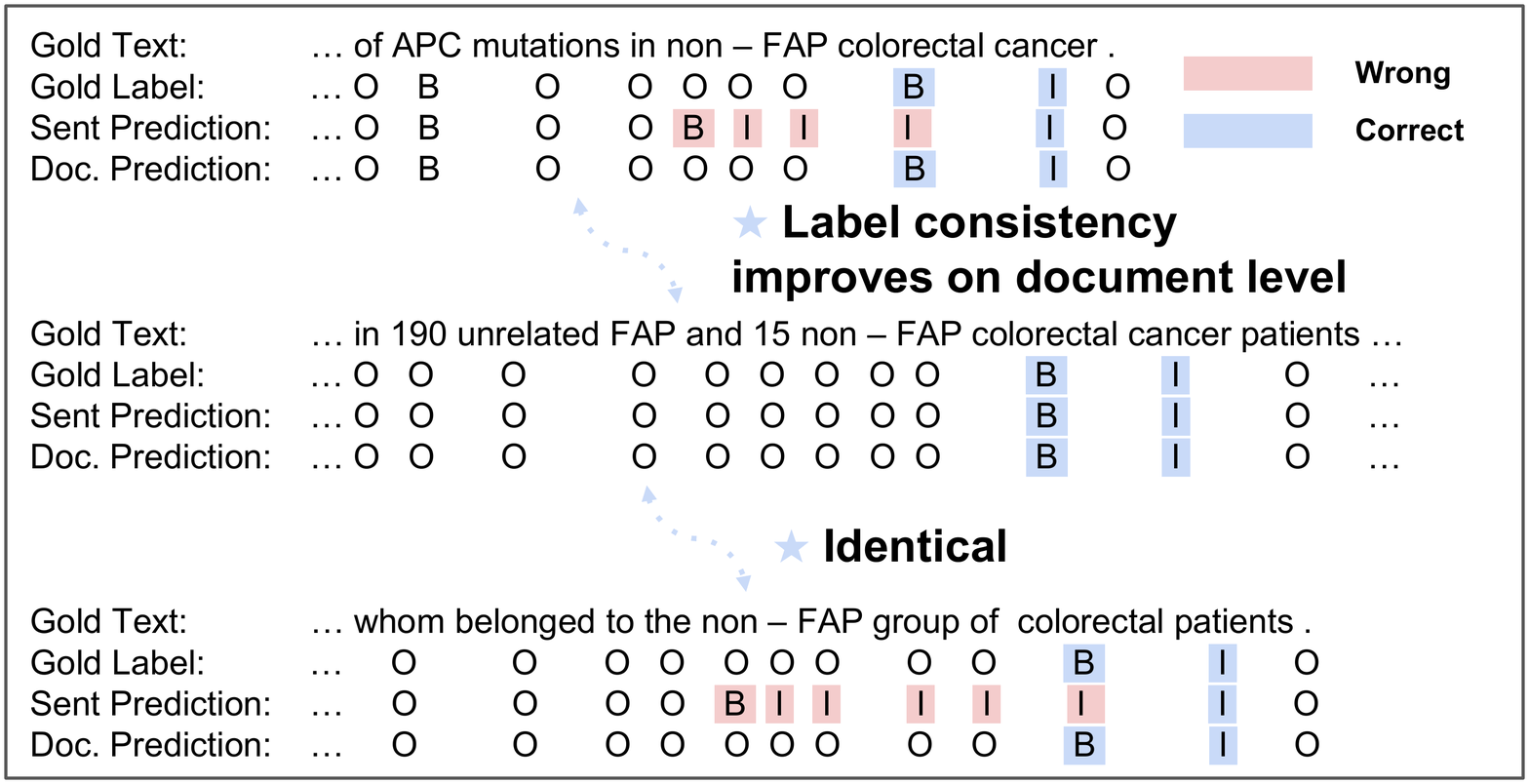}
  \includegraphics[width=0.35\textwidth]{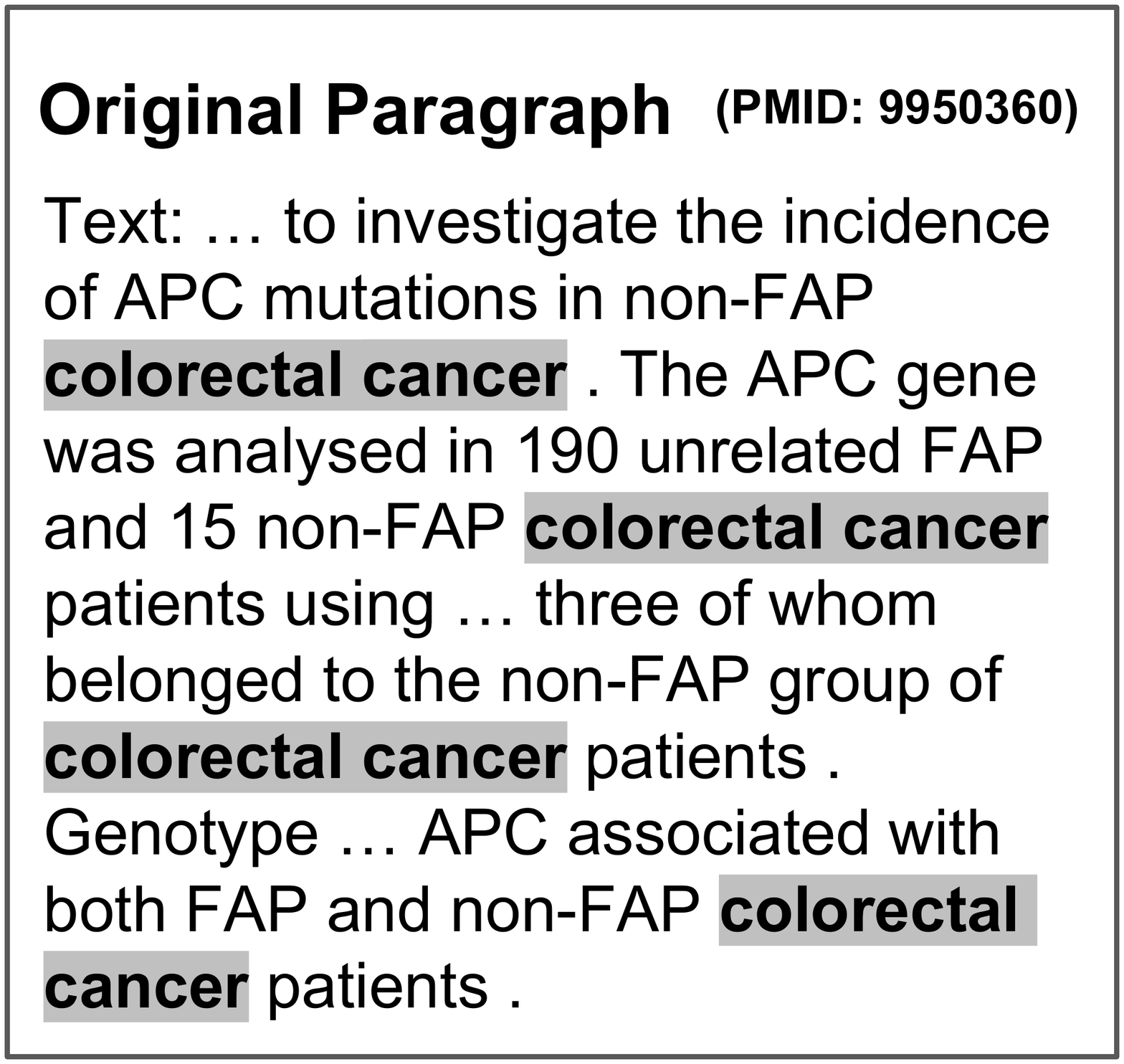}
  \vspace{-1.25cm}
  \caption{
  Overview of motivating examples of entity-aware attributes. (Left) We provide an example with a gold label, sentence prediction, and document prediction. Most of the biomedical entities contain the adjective or preposition within the entities. Suggesting a sufficient context improves label agreement of those adjectives such as \textbf{`colorectal'}. (Right) It is worth noting that the majority of biomedical datasets are derived from its golden paragraph (PubMed).
  }\label{Fig:conner_overview}
  \vspace{-0.25cm}
\end{figure*}

\begin{figure}[t]
    \centering
    \includegraphics[width=0.235\textwidth]{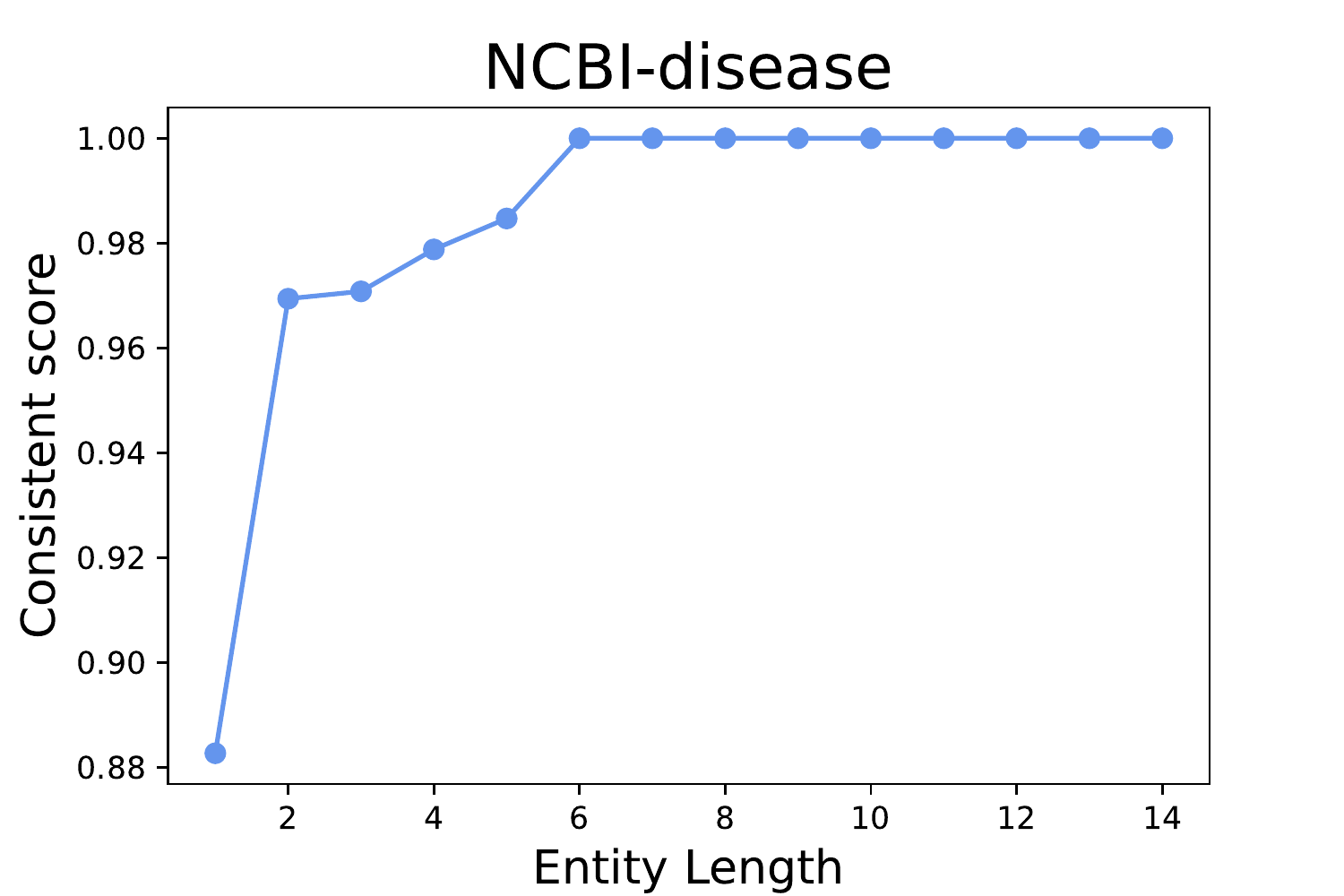}
    \includegraphics[width=0.235\textwidth]{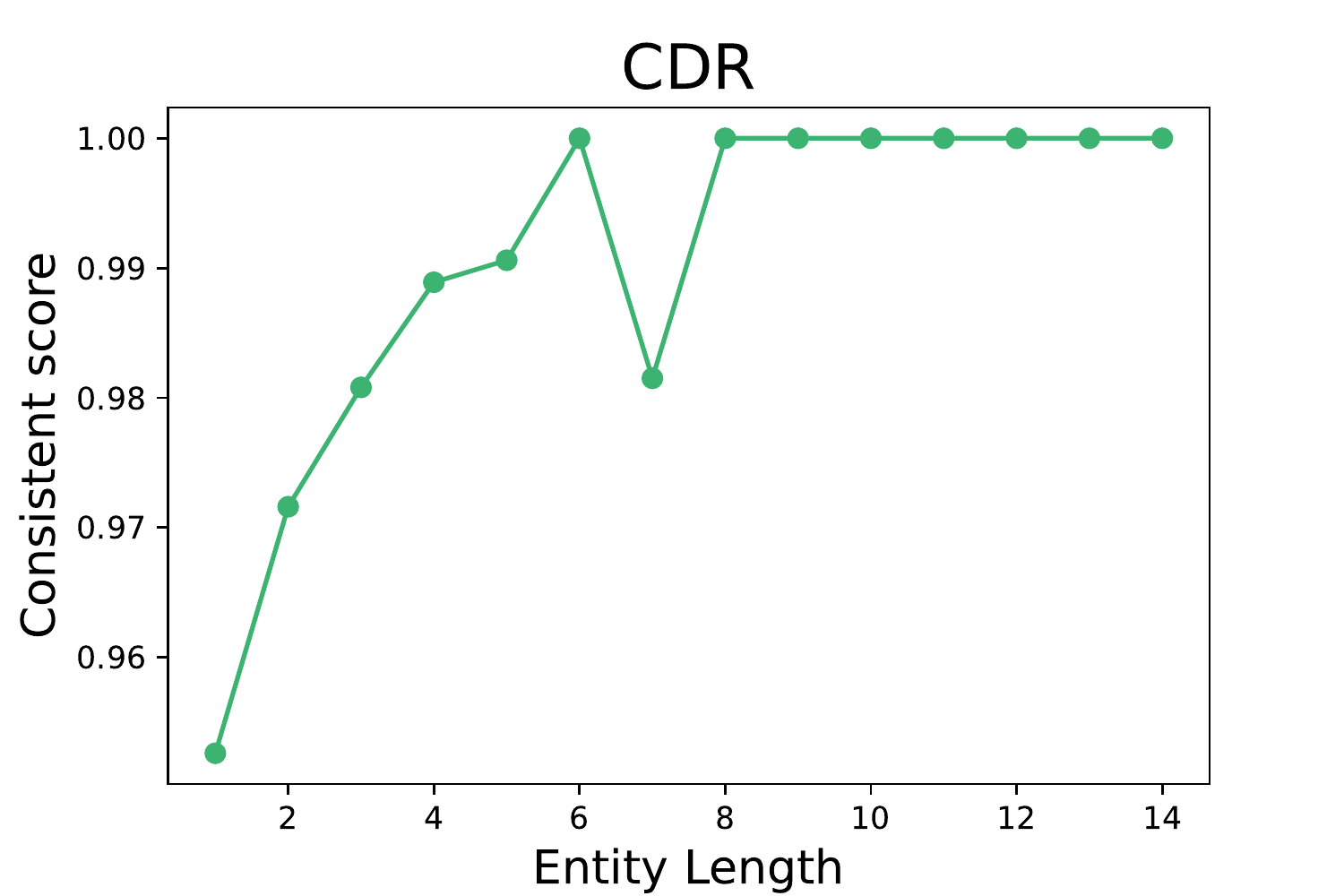}  \\ \includegraphics[width=0.235\textwidth]{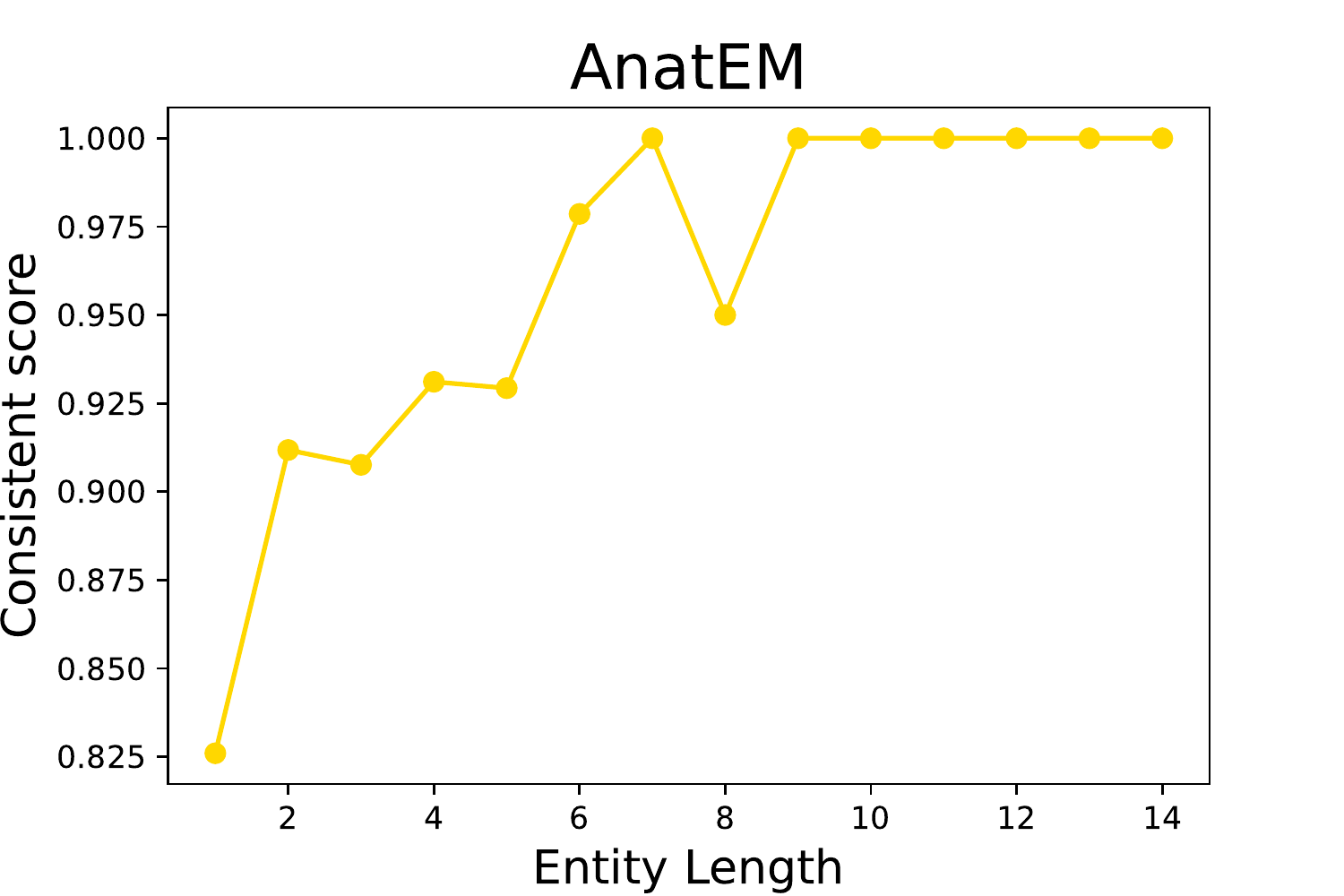}  \includegraphics[width=0.235\textwidth]{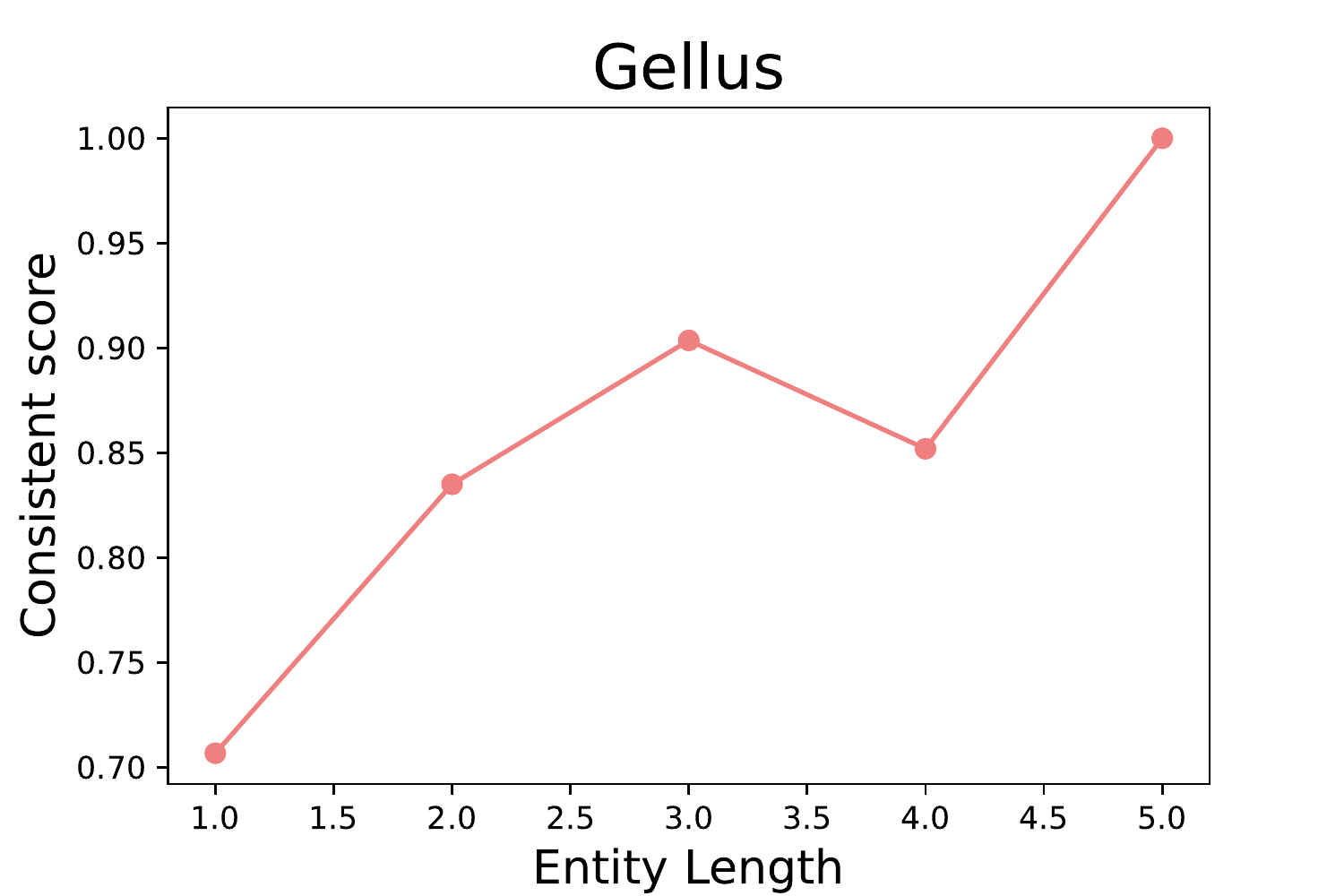}
    \caption{
    Consistent score per entity length. The x-axis denotes the entity length (i.e., $\phi_{eLen}$), and the y-axis denotes the consistent score (i.e., $\phi_{eCon}$). Short lengths of entities show low consistent scores due to the low label agreement of the modifier tokens.
    }
    \label{Fig:consistent}
\end{figure}

We first clarify why we need the NER model to make consistent predictions. 
We provide our motivating example in~\Cref{Fig:conner_overview}.
For example, the mention `\textbf{colorectal cancer}' is an entity of the disease type.
Predicting such a mention is challenging in a sentence context owing to context incompleteness.
As a result, the sentence model produces an error in predicting `\textbf{non – FAP}' or `\textbf{colorectal}'.
Meanwhile, the NER model trained on document contexts shows consistent predictions because the token `\textbf{colorectal}' occurs frequently within the documents.
Therefore, the document provides sufficient label agreement to learn label representations of `\textbf{colorectal}'.
Although the document NER model shows consistent predictions and much better performance than the sentence NER model (\Cref{tab:ner_results}), it still falls behind our expectations.
Models trained on document contexts continue to produce 64\% errors that contain modifiers (i.e., prepositions or adjectives), such as \textit{primary, hereditary,} and \textit{congenital} (\Cref{tab:analysis}).
Because modifiers are used as both entity and non-entity tokens depending on the context situation, they are difficult to predict.
Therefore, the modifiers exhibit a low label consistency score and occur at short entity lengths (\Cref{Fig:consistent}).

\begin{table}[t]
\processtable{Results on four biomedical NER benchmarks. We use BioBERT~\citep{lee2020biobert} and BioLM~\citep{lewis-etal-2020-pretrained} to compare the predictions between different context levels. We report the F1 score to evaluate the models.
\label{tab:ner_results}} {\resizebox{1.0\columnwidth}{!}{{
\begin{tabular}{ c l ccccc }
\toprule
\multicolumn{1}{c }{\multirow{2}{*}{Context}} & \multicolumn{1}{c}{\multirow{2}{*}{Model}} & \multicolumn{5}{c }{Evaluation (F1)}                                                                 \\ \cmidrule{3-7} 
\multicolumn{1}{c}{}                               &                        & \multicolumn{1}{c}{NCBI-disease} & \multicolumn{1}{c}{CDR}  & \multicolumn{1}{c}{AnatEM} & Gellus & \multicolumn{1}{c}{Avg.} \\ \midrule
\multirow{2}{*}{Sentence}                            & BioBERT                & \multicolumn{1}{c}{89.0}         & \multicolumn{1}{c}{89.1} & \multicolumn{1}{c}{73.9}      & \multicolumn{1}{c}{54.9} & \multicolumn{1}{c}{76.7}      \\ 
& BioLM                  & \multicolumn{1}{c}{88.3}         & \multicolumn{1}{c}{89.5} & \multicolumn{1}{c}{74.9}      & \multicolumn{1}{c}{55.9}     & \multicolumn{1}{c}{77.2}     \\ \midrule
\multirow{2}{*}{Document}                            & BioBERT                & \multicolumn{1}{c}{89.1}         & \multicolumn{1}{c}{89.9} & \multicolumn{1}{c}{77.1}   & \multicolumn{1}{c}{56.3}   & \multicolumn{1}{c}{78.1}    \\ 
& BioLM                  & \multicolumn{1}{c}{88.5}         & \multicolumn{1}{c}{90.2} & \multicolumn{1}{c}{80.3}   & \multicolumn{1}{c}{56.9}   & \multicolumn{1}{c}{79.0}    \\
\bottomrule
\end{tabular}}}}{
}
\vspace{-0.5cm}
\end{table}

To avoid the aforementioned errors, we present  \textit{ConNER} which enhances the label dependency of modifiers to achieve higher label agreement.
An abstract of biomedical literature is fed into the biomedical pre-trained language model, BioBERT~\citep{lee2020biobert} or BioLM~\citep{lewis-etal-2020-pretrained}, to output context representation (\Cref{sec:document_ner}).
On top of the pre-trained language model, we propose label refinement to improve the label representations of uncertain tokens within entities (\Cref{sec:label_refinement}).
We also suggest our loss term for biomedical entities to resemble the label representations on two different architectures: fully connected layers (MLP) and bidirectional long-short term memory (BiLSTM) architecture.
We use MLP as the main classification layer to generate final output representations of raw text and use the BiLSTM architecture to generate label representations of biomedical entities.
We adopt the notion that the MLP layer is more robust to long entities~\citep{lafferty2001conditional}, which is an essential property of biomedical entities.
We employ the BiLSTM architecture to improve the label dependency of biomedical entities~\citep{collobert2011natural, lample2016neural}.
To demonstrate the effectiveness of the proposed ConNER approach, we use four biomedical benchmarks.
On three datasets, we achieve a higher F1 score than previous state-of-the-art models~\citep{wang2021improving,lewis-etal-2020-pretrained}.
In particular, our model achieves higher label agreement and proves its efficacy on two datasets with 7.5–8.6\% absolute improvements in the F1 score (\Cref{sec:experimental_results}).
We provide an interpretation of the effectiveness of our method for a dataset that intrinsically has a low label consistency score (\Cref{sec:qualitative_analysis}).

The contributions of this study are summarized as follows.
(1) We investigate why document NER models make inconsistent predictions in biomedical domains.
Based on our observations, we observe that modifiers (i.e., adjectives and prepositions) exhibit a low label consistency score and produce errors by making inconsistent predictions.
(2) We present our method, \textit{ConNER}, which enhances the label dependency of modifiers to generate improved label representations.
(3) The results of the experiments show that our ConNER approach significantly improves the accuracy of document NER models, indicating that it can help achieve the highest level of label agreement.
(4) For different tasks related to low label consistency, we show that ConNER outperforms various baselines, and we analyze the factors influencing its performance.

\section{Background}
\subsection{Named Entity Recognition}
\label{document_ner}
The goal of NER is to find a word or phrase that corresponds to a specific instance, such as a person, location, organization, or any other miscellaneous entity.
The NER task primarily involves the extraction and classification of named entities found in a corpus with pre-defined entity tags.
We use BIO tagging~\citep{ramshaw1999text}, where a B- prefix (Beginning) indicates the beginning of the chunk following an I- prefix (Inside), and an O- prefix (Outside) indicates being inside and not belonging to the chunk.
The task commonly uses two different decoding strategies: tag-independent decoding (i.e., MLP~\citep{collobert2011natural}) and tag-dependent decoding (i.e., CRF~\citep{lafferty2001conditional, lample2016neural}).
Here, we use only the tag-independent decoding strategy.

\subsection{Attribute Definition}
\label{attribue_definition}
Following previous works~\citep{fu2020interpretable, fu2021larger}, we define the term \textit{attribute} as a value that characterizes the properties of an entity that may be correlated with performance.
The authors introduced attributes bridging the gap between the final performance (we use the F1 score) and interpretable evaluation based on model predictions.
Assuming that one attribute is given, the test set of NER tasks naturally partitions into several interpretable buckets, we investigate whether the attribute affects the final performance bucket-wise.

\begin{figure}[t]
  \centering
  \includegraphics[width=0.5\textwidth]{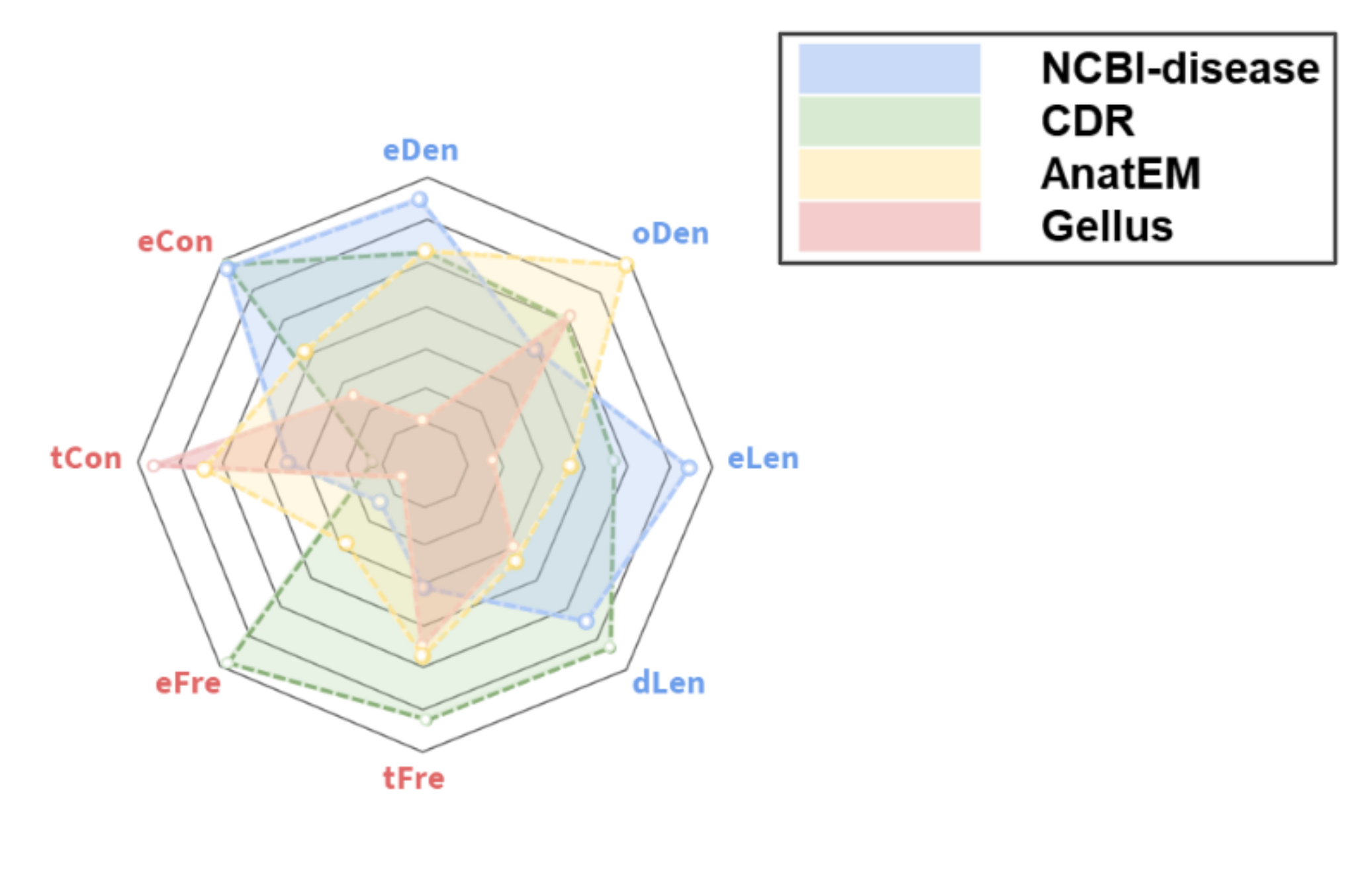}
  \caption{
  Overview of Dataset Biases. Each dot denotes an attribute score of entities. Intuitively, these intrinsic differences in datasets explain what factors have a significant influence on improving performance.
  }\label{Fig:dataset_bias}
  \vspace{-0.5cm}
\end{figure}

Formally, we define notations to facilitate the definition of attributes.
Given a set of documents $D$, entity tagging aims to extract a set of entities $\mathcal{E}$ as spans or tokens. 
We first denote a span set $\mathcal{E}^{tr}$ $\subset$ $\mathcal{E}$ as an argument.
Specifically, \citep{fu2020interpretable} introduced a feature function $\phi(\cdot)$ to aggregate features to interpret the properties of tagged entities $\mathcal{E}^{tr}$:
\begin{equation}
    \mathcal{F}(\textbf{x}, \phi(\cdot), D(\mathcal{E})) = \frac{|\{\varepsilon | \phi(\varepsilon) = \phi(\textbf{x}), \forall \varepsilon \in D(\mathcal{E})\}|}{|D(\mathcal{E})|},
\end{equation}
where $\textbf{x}$ denotes an entity span that can also replace it as token $x$.
Similarly, we define some training set-independent attribute functions as follows:
\begin{equation}
    \phi_{tLen}(x) = |x|: \text{token span length}
\end{equation}
\begin{equation}
    \phi_{eLen}(\textbf{x}) = |\textbf{x}|: \text{entity span length}
\end{equation}
\begin{equation}
    \phi_{dLen}(\textbf{x}) = |doc(\textbf{x})|: \text{document length}
\end{equation}
\begin{equation}
    \phi_{eDen}(\textbf{x}) = |ent(doc(\textbf{x}))|/\phi_{dLen}(\textbf{x}): \text{entity density}
\end{equation}
\begin{equation}
    \phi_{oDen}(\textbf{x}) = |oov(doc(\textbf{x}))|/\phi_{dLen}(\textbf{x}): \text{OOV density}
\end{equation}
Rather than using a sentence-level context, we provide a document-length attribute based on the usage of the document-level context.
We use two functions to define the density of entity $ent(\cdot)$, which counts the number of entity words, and the density of out-of-vocabulary words $oov(\cdot)$, which counts the number out of training set words.
We also leverage the training set-dependent attribute functions, as follows:
\begin{equation}
    \phi_{tFre}(x) := \mathcal{F}(x, \phi_{str}(\cdot), D(\mathcal{E}^{tr})): \text{token frequency}
\end{equation}
\begin{equation}
    \phi_{eFre}(\textbf{x}) := \mathcal{F}(\textbf{x}, \phi_{str}(\cdot), D(\mathcal{E}^{tr})): \text{entity frequency}
\end{equation}
\begin{equation}
    \phi_{tCon}(x) := \mathcal{F}(x, \phi_{label}(\cdot), D(\mathcal{E}^{tr})): \text{label consistency of token}
\end{equation}
\begin{equation}
    \phi_{eCon}(\textbf{x}) := \mathcal{F}(\textbf{x}, \phi_{label}(\cdot), D(\mathcal{E}^{tr})): \text{label consistency of entity}
\end{equation}
where $\phi_{str}(\cdot)$ and $\phi_{label}(\cdot)$ denote a string and label of the corresponding argument, respectively, and 
$\phi_{tCon}(x)$ and $\phi_{eCon}(\textbf{x})$ refer to measuring how consistently a certain token or entity span is assigned to a pre-defined label, respectively.
Here, we define \textit{label consistency} as the degree of label agreement of n tokens/entities in the training set.
We use attribute functions to interpret which dataset attributes have an impact on performance improvement~(\Cref{Fig:dataset_bias}).

\section{Method}

\begin{figure*}[t]
\vspace{-1cm}
  \centering
  \includegraphics[width=\textwidth]{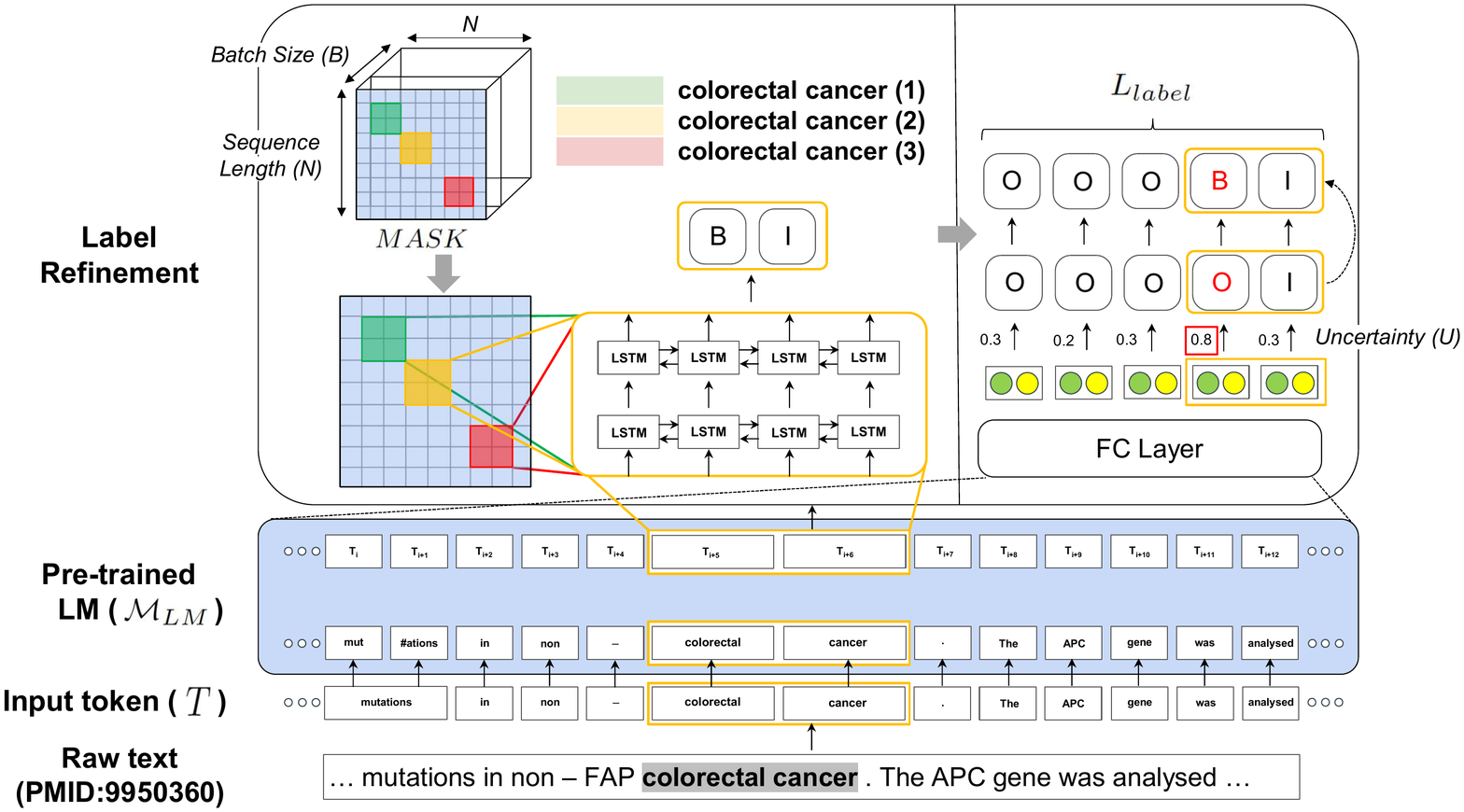}
  \vspace{-1.5cm}
  \caption{
  Overview of~\ours~performing label refinement on biomedical entities. The abstract of biomedical literature is fed into the pre-trained language model $\mathcal{M}_{LM}$ to output context representation. In the label-refinement process, we use a mask tensor $MASK$ to find the position of biomedical entities and feed it into the BiLSTM architecture $\mathcal{M}_{LSTM}$ to improve label dependency of modifiers within entities that induce low label agreement. On top of the fully connected layer, we compute an uncertainty score $U$ to determine which tokens need to be modified for their predicted label.
  }\label{Fig:conner_structure}
\end{figure*}

In this section, we start by learning the document-level model through the supervision of named entity recognition tasks.
Our goal is to learn how to enrich the label representations of biomedical entities by improving their label consistency~(\Cref{sec:document_ner}).
We then introduce our label refinement on biomedical entities, which encourages label representations of uncertain tokens within entities~(\Cref{sec:label_refinement}).
Finally, we discuss the use of a loss term for biomedical entities to resemble the label representations of two different architectures.
~\Cref{Fig:conner_structure} depicts the overall structure of~\ours~approach.

\subsection{Document Named Entity Recognition}
\label{sec:document_ner}
\ours~consists of a pre-trained language model $\mathcal{M}_{LM}$ and label-refinement process, as shown in~\Cref{Fig:conner_structure}.
For document tagging, we use abstracts of biomedical literature as raw input and make predictions for each word.
Let $D_i = \{ T_{i1}, T_{i2}, ..., T_{iN} \} $ represent a sequence of tokens, where $N$ denotes the total number of tokens in a document $D_i$.
First, we apply a biomedical pre-trained language model (e.g., BioBERT~\citep{lee2020biobert} or BioLM~\citep{lewis-etal-2020-pretrained}) to obtain contextualized word representations for each token: $T_{i1}, T_{i2}, ..., T_{iN} \in \mathbb{R}^d$. 
Then, we represent a biomedical entity $\textbf{e}$ as the concatenation of start-to-end vector representations, as follows:
\begin{equation}
    \mathcal{M}_{LM}(e) = [T_{start}:T_{end}] \in \mathbb{R}^{\phi_{eLen}(\textbf{e}) \cdot d}
\end{equation}
where $T_{start}$ and $T_{end}$ denote the start and end of biomedical entity $\textbf{e} \in \mathcal{E}$, respectively, and $d$ denotes the final hidden dimension of language model $\mathcal{M}_{LM}$.
Note that we use the notation $\textbf{e}$ as an example to help understand our method \ours.
We use tag-independent decoding (i.e., MLP) on the model $\mathcal{M}_{LM}$ as the main classification layer, which is useful for predicting long-named entities~\citep{fu2020interpretable}.
Formally, we can define our classification loss $L_{class}$ using cross-entropy objectives to optimize our model as follows:
\begin{equation}
\label{eq:class}
    L_{class} = -\frac{1}{N}\sum_{j=1}^{N}y_{j} \cdot log(p_{j})
\end{equation}
where $y_j$ denotes the ground-truth label and $p_j$ denotes the probability of the main classification layer.

\paragraph{Sentence vs. Document.}
In our pilot experiments, we first observe that the label consistency of entity $\phi_{eCon}(\textbf{x})$ is enhanced when we expand the context from sentences to documents.
We also observe that using a larger context is helpful in achieving better performance (+1.4-1.8\% in F1 score), as shown in~\Cref{tab:ner_results}.
Based on our positive observations, we hypothesize that one key reason for the performance gain is that biomedical entities are challenging to predict at the sentence level owing to context incompleteness.
Learning from the limited context makes it difficult to predict a token within an entity such as `\textbf{colorectal}' (see ~\Cref{Fig:conner_overview}).

\subsection{Label Refinement on Biomedical Entities}
\label{sec:label_refinement}
Overall, we propose a label-refinement process that depends on an entity-length attribute $\phi_{eLen}(\textbf{e})$ to enrich the label representations of the entity spans.
To encourage label representations, we add a loss term of the entity span to make the label distribution similar between the main classification layer and the refinement process.
We use the MLP layer to obtain a final output representation and both an uncertainty and a draft label for each token representation.
We use a BiLSTM architecture~\citep{lafferty2001conditional} to improve the label dependency of tokens within biomedical entities and to generate label representations of entities.
These label representations from the BiLSTM architecture provide a precise label to the main classification layer by verifying the uncertainty of the draft label (see~\Cref{Fig:conner_overview}).

The independent decoder predicts well for longer entities, especially biomedical entities~\citep{fu2020interpretable, fu2021larger, jeong2021regularization}.
With the preceding observation, we find that the modifiers (i.e., prepositions or adjectives) are both used as entity and non-entity tokens.
We also find that a low label consistency occurs at short lengths of entities (i.e., $\phi_{eLen}(\textbf{e}) < 5$), which most modifiers consist of in this scope.
As a result, we attempted to deal with long entities by employing the MLP layer and developing a label-refinement process to target those modifiers to be predicted consistently.

\paragraph{Uncertainty.}
We ask the natural question: \textit{How can we decide what tokens should be refined?}
We choose this criterion for the certainty of the draft label on token representations~\citep{gui2021leveraging}.
We calculate the uncertainty $U$ of the probability distributions of the MLP layer as follows:
\begin{equation}
\label{eq:uncertainty}
    U = H(p_{c}) = -\sum_{c=1}^{C}p_{c}~\text{log}~p_{c}.
\end{equation}
where $C$ denotes the number of pre-defined entity types per dataset.
We use the entropy of the probability distribution $p_{c}$ to define the uncertainty $U$.
Given this uncertainty score, label refinement proposes a positive training signal to shift the draft label.

\paragraph{Label Refinement.}
We decide to improve the label dependency on modifiers (i.e., adjectives or prepositions) used in the entities during training.
Specifically, we create an entity-aware mask tensor $MASK \in \mathcal{R}^{B \cdot N \cdot N}$ 
to indicate the position of the padded indices such that our model does not attend to non-entities, where $B$ denotes the batch size.
The mask tensor is applied to the BiLSTM layers $\mathcal{M}_{LSTM}$ to obtain a label representation ${l}(\textbf{e})$, as follows:
\begin{equation}
\label{eq:refine}
    l(\textbf{e}) = \mathcal{M}_{LSTM}\{MASK(\mathcal{M}_{LM}(\textbf{e}))\} \in \mathbb{R}^{\phi_{eLen}(\textbf{e}) \cdot C}.
\end{equation}

Then, $\mathcal{M}_{LSTM}$ trains to softly assist the main classification layer in performing precise predictions with improved consistency.
Note that we use the term \textit{softly assist} to describe a process that assists the main classification layer in shifting its draft label during training.
That is, the trained model $\mathcal{M}_{LSTM}$ enhances the label dependency of an uncertain token and softly assists in refining the draft label of the uncertain token.
Formally, we add element-wisely to assist predictions before calculating a label loss $L_{label}$, as shown below:
\begin{equation}
\label{eq:oplus}
    p = p \oplus l,
\end{equation}
\begin{equation}
    L_{label} = -\frac{1}{N}\sum_{j=1}^{N}y_{j} \cdot log(l_{j})
\end{equation}
where $y_j$ and $l_j$ denote the ground-truth label and probability of the $\mathcal{M}_{LSTM}$ layer, respectively, and $p$ is used to compute the classification loss in~\Cref{eq:class}.
Finally, we determine the criterion using the pre-computed uncertainty score $U$ in~\Cref{eq:uncertainty}. 
We set an uncertainty threshold $\Gamma$ to distinguish the tokens that should be refined within entities.
For example, in~\Cref{Fig:conner_structure}, a token `\textbf{colorectal}' of the entity `\textbf{colorectal cancer}' is first predicted as an Outside tag.
During training, the token `\textbf{colorectal}' is converted to a Beginning tag because of the high uncertainty score.
We found that $\Gamma=0.3$ worked well in practice.
See~\Cref{Fig:threshold} for an ablation study.

\paragraph{Distillation.}
\label{sec:distill}
We propose improving the label representations of biomedical entities by distilling knowledge~\citep{hinton2015distilling} from our decoding layers. 
We minimize the Kullback-Leibler divergence between the probability distribution from the tag-independent layer (i.e., MLP) and the tag-dependent layer (i.e., BiLSTM) on top of the biomedical pre-trained language model. 
The distillation loss was computed as follows:

\begin{equation}
\label{eq:distill}
    L_{distill} = \frac{KL(p||l) + KL(l||p)}{2},
\end{equation}
where $p$ and $l$ denote the probability distributions of the MLP and BiLSTM layers, respectively.
Note that distillation is computed before~\Cref{eq:oplus}.

\subsection{Training Objective}
\label{training_objective}
We optimize the three losses altogether to improve the label agreement of the entity through label refinement and distilling knowledge while predicting on the tag-independent layer.
Our final loss is computed as follows:
\begin{equation}
    L = \lambda_{1}L_{class} + \lambda_{2}L_{label} + \lambda_{3}L_{distill}
\end{equation}
where $\lambda_{1}, \lambda_{2}, \lambda_{3}$ scale the importance of each loss term.
We observe that $\lambda_{1} = 1$, $\lambda_{2}=1e-1$, and $\lambda_{3}=1e-3$ exhibit the best performance in our framework.
See~\Cref{tab:ablation} for an ablation study of the other components.

\section{Experimental Setup}
\begin{table}[t]
\processtable{
Statistics of four biomedical named entity recognition datasets. The CDR dataset suggests two entity types that share the same text. Annotations (Ann.) and Unique Ann. refer to the entire entity and its unique numbers respectively.
\label{tab:dataset_stats}
}
{\resizebox{1.0\columnwidth}{!}{
\begin{tabular}{ l l c c c c c } 
\toprule
Corpora      & Entity Type                                                  & Documents & Sentences & Tokens & Annotations & Unique Ann. \\
\midrule
NCBI-disease & Disease & 793 & 7,142 & 180,992 & 6,881 & 2,502 \\
CDR          & \begin{tabular}[c]{@{}l@{}}Disease /\\ Chemical\end{tabular} & 1,500 & 14,503 & 346,019 & \begin{tabular}[c]{@{}l@{}} 12,957 / \\ 15,837 \end{tabular} & \begin{tabular}[c]{@{}l@{}} 4,477 / \\ 3,765 \end{tabular} \\
AnatEM       & Anatomic & 1,212 & 11,809 & 298,780 & 13,692 & 5,042 \\
Gellus       & Cell Lines & 1,212 & 11,809 & 312,584 & 650 & 243 \\
\bottomrule
\end{tabular}}}{}
\vspace{-0.2cm}
\end{table}

\begin{table}[t]
\processtable{Hyperparameters of the four biomedical named entity recognition settings. For common hyperparameter settings, we focus on searching batch size, learning rate, and training epoch for learning schemes. We also focus on searching for the optimal hyperparameters of the uncertainty threshold $\Gamma$.
\label{tab:hyperparameter}} {\resizebox{1.0\columnwidth}{!}{{
\begin{tabular}{ c l l c c c c c c c}
\toprule
\multirow{2}{*}{Context}  & \multicolumn{1}{c}{\multirow{2}{*}{Corpora}} & \multicolumn{1}{c}{\multirow{2}{*}{\begin{tabular}[c]{@{}c@{}}Pretrained\\ LM\end{tabular}}} & \multirow{2}{*}{\begin{tabular}[c]{@{}c@{}}Batch\\ Size\end{tabular}} & \multirow{2}{*}{\begin{tabular}[c]{@{}c@{}}Learning\\ Rate\end{tabular}} & \multirow{2}{*}{\begin{tabular}[c]{@{}c@{}}Training\\ Epoch\end{tabular}} & \multirow{2}{*}{\begin{tabular}[c]{@{}c@{}}Sequence\\ Length\end{tabular}} & \multirow{2}{*}{\begin{tabular}[c]{@{}c@{}}Threshold\\$\Gamma$\end{tabular}} & \multirow{2}{*}{$\lambda_{1}$} & \multirow{2}{*}{$\lambda_{2}$} \\
 & \multicolumn{1}{c}{} & \multicolumn{1}{c}{} & & & & & & & \\ \midrule
\multirow{4}{*}{Document} & NCBI-disease & BioLM & 6 & 3e-5 & 50 & 512 & 0.3 & 1e-1 & 1e-3 \\ 
 & CDR & BioLM & 6 & 3e-5 & 30 & 512 & 0.3 & 1e-1 & 1e-3 \\
 & AnatEM & BioLM & 6 & 5e-5 & 30 & 512 & 0.3 & 1e-1 & 1e-3 \\
 & Gellus & BioLM & 6 & 3e-5 & 40 & 512 & 0.3 & 1.0 & 1e-3 \\ \bottomrule
\end{tabular}}}}{
}
\vspace{-0.5cm}
\end{table}

\subsection{Dataset}
\label{dataset}
We use four biomedical NER benchmarks across four entity types: NCBI-disease~\citep{dougan2014ncbi}, CDR~\citep{li2016biocreative}, AnatEM~\citep{pyysalo2014anatomical}, and Gellus~\citep{kaewphan2016cell}, following the standard train/dev/test splits for biomedical NER evaluation. 
(1) \textbf{NCBI-disease}~\citep{dougan2014ncbi} consists of 793 PubMed abstracts with manually annotated disease entities. 
(2) \textbf{CDR}~\citep{li2016biocreative} contains 1,500 PubMed abstracts manually annotated with disease and chemical entities in the same context.
(3) \textbf{AnatEM}~\citep{pyysalo2014anatomical} consists of 1,212 PubMed abstract and full-text extracts annotated with 12 anatomical entity types.
(4) \textbf{Gellus}~\citep{kaewphan2016cell} consists of annotating cell lines in 1,212 documents from PubMed abstracts and PMC full-text extracts. Half of the corpora were drawn from the AnEM corpus~\citep{ohta2012open}, and the other half were drawn from the BioNLP ST'13 Cancer Genetics (CG) task documents~\cite{pyysalo2013overview}.
\Cref{tab:dataset_stats} shows the dataset statistics.

\begin{table*}[t]
\processtable{Results on biomedical NER benchmarks. F1 score is reported. The best score is displayed in bold and the second-best score is underlined. $^{\dagger}$ numbers are estimated from the figures in the original papers.
\label{tab:conner_results}} {\resizebox{0.9\textwidth}{!}{{
\begin{tabular}{ c l cccc }
\toprule
\multirow{2}{*}{Context} & \multicolumn{1}{c}{\multirow{2}{*}{Model}} & \multicolumn{4}{c}{Evaluation (F1)} \\ \cmidrule{3-6} 
&                        & \multicolumn{1}{c}{NCBI-disease} & \multicolumn{1}{c}{CDR}  & \multicolumn{1}{c}{AnatEM} & Gellus \\ \midrule
\multirow{6}{*}{Sentence}        
& B-MTM~\citep{crichton2017neural}$^{\dagger}$                  & \multicolumn{1}{c}{80.4}         & \multicolumn{1}{c}{89.2} & \multicolumn{1}{c}{82.2}   & -      \\ 
& BiLSTM-CRF~\citep{habibi2017deep}$^{\dagger}$             & \multicolumn{1}{c}{84.6}         & \multicolumn{1}{c}{-}    & \multicolumn{1}{c}{-}      & \textbf{75.6}   \\ 
& BioBERT~\citep{lee2020biobert}                & \multicolumn{1}{c}{89.0}         & \multicolumn{1}{c}{89.1} & \multicolumn{1}{c}{73.9}   & 54.9   \\ 
& BioLM~\citep{lewis-etal-2020-pretrained}                  & \multicolumn{1}{c}{88.3}         & \multicolumn{1}{c}{89.5} & \multicolumn{1}{c}{74.9}   & 55.9   \\
& CL-L2 (w/o context)~\citep{wang2021improving}$^{\dagger}$    & \multicolumn{1}{c}{\underline{89.2}}         & \multicolumn{1}{c}{90.7} & \multicolumn{1}{c}{-}      & -      \\  
& CL-KL (w/o context)~\citep{wang2021improving}$^{\dagger}$    & \multicolumn{1}{c}{\underline{89.2}}         & \multicolumn{1}{c}{90.7} & \multicolumn{1}{c}{-}      & -      \\ 
\midrule
\multirow{3}{*}{Document}      
& CL-L2 (w/ context)~\citep{wang2021improving}$^{\dagger}$     & \multicolumn{1}{c}{\underline{89.2}}         & \multicolumn{1}{c}{\underline{91.0}} & \multicolumn{1}{c}{-}      & -      \\ 
& CL-KL (w/ context)~\citep{wang2021improving}$^{\dagger}$     & \multicolumn{1}{c}{89.0}         & \multicolumn{1}{c}{90.9} & \multicolumn{1}{c}{-}      & -      \\ 
& ConNER (Ours)                &
\multicolumn{1}{c}{\textbf{89.9}}         & \multicolumn{1}{c}{\textbf{91.3}} & \multicolumn{1}{c}{\textbf{83.5}}   & \underline{63.4}   \\
\bottomrule
\end{tabular}}}}{ 
} 
\end{table*}

\begin{table*}[t]
\processtable{An ablation study of~\ours~components. We perform three different experiments: removing distillation (—~\{$L_{distill}$\}), label refinement (—~\{$L_{label}$\}), and both of these process (—~\{$L_{distill}$,$L_{label}$\}). We used our best hyperparameter setting on the BioLM~\citep{lewis-etal-2020-pretrained} model equally. The best scores are displayed in bold.
\label{tab:ablation}} {\resizebox{0.9\textwidth}{!}{{
\begin{tabular}{lcccccccccccc}
\toprule
\multicolumn{1}{c}{\multirow{2}{*}{Model Description}} & \multicolumn{12}{c}{Evaluation (P/R/F1)} \\ \cmidrule{2-13} 
\multicolumn{1}{c}{}                             & \multicolumn{3}{c}{NCBI-disease} & \multicolumn{3}{c}{CDR} & \multicolumn{3}{c}{AnatEM} & \multicolumn{3}{c}{Gellus} \\ \midrule
\ours                                             &  \multicolumn{3}{c}{\textbf{88.8} / \textbf{91.1} / \textbf{89.9}}             & \multicolumn{3}{c}{\textbf{89.9} / \textbf{92.7} / \textbf{91.3}}    & \multicolumn{3}{c}{\textbf{83.2} / \textbf{83.7} / \textbf{83.5}}       & 
\multicolumn{3}{c}{\textbf{62.6} / \textbf{64.2} / \textbf{63.4}} \\ 
\ours~—~\{$L_{distill}$\}                              & \multicolumn{3}{c}{87.5 / 89.9 / 88.7}             & \multicolumn{3}{c}{89.2 / 92.3 / 90.7}    & \multicolumn{3}{c}{81.6 / 81.1 / 81.3}       &   \multicolumn{3}{c}{54.3 / 60.3 / 57.1}     \\ 
\ours~—~\{$L_{label}$\}                                & \multicolumn{3}{c}{87.4 / 90.4 / 88.8}             & \multicolumn{3}{c}{89.4 / 92.1 / 90.7}    & \multicolumn{3}{c}{81.0 / 81.9 / 81.5}      &    \multicolumn{3}{c}{54.2 / 61.2 / 57.5}    \\
\ours~—~\{$L_{distill}$,$L_{label}$\}                 & \multicolumn{3}{c}{87.4 / 89.7 / 88.5}             & \multicolumn{3}{c}{89.8 / 90.7 / 90.2}    & \multicolumn{3}{c}{80.3 / 80.4 / 80.3}       & \multicolumn{3}{c}{54.8 / 59.1 / 56.9} \\ \bottomrule
\end{tabular}}}}{
}
\vspace{-0.5cm}
\end{table*}

\subsection{Comparison Methods}
We evaluate the sentence- and document-level contexts and compare ~\ours~ with several neural network models commonly used in biomedical domains.
(1) \textbf{B-MTM}~\citep{crichton2017neural} developed a multi-task learning model using various biomedical sources annotated with different entity types.
(2) \textbf{BiLSTM-CRF}~\citep{habibi2017deep} proposed a combination of word embeddings and LSTM with the CRF decoding strategy in the biomedical domain.
(3) \textbf{BioBERT}~\citep{lee2020biobert} introduced a biomedical-specific language representation model pre-trained on large-scale biomedical corpora.
(4) \textbf{BioLM}~\citep{lewis-etal-2020-pretrained} suggested a biomedical-specific language representation model pre-trained on biomedical and clinical corpora.
(5) \textbf{CL-KL and CL-L2}~\citep{wang2021improving} proposed a method that can retrieve and select a semantically relevant context using a search engine to improve contextual representations, with the original sentence as a query.

\subsection{Implementation Details}
We train~\ours~using BioLM~\citep{lewis2020pretrained} or BioBERT~\citep{lee2020biobert} to treat biomedical entity types.
These two pre-trained language models are commonly used as backbone models in the biomedical domain, and we adopt these models to generate contextualized representations.
We set 128 tokens as the maximum sequence length for the sentence-level context and 512 tokens for the document-level context, and tokens with more than the maximum sequence length were truncated.
The batch size was set to 32 for the sentence level and 6 for the document level.
We select a learning rate in the range \{3e-5, 5e-5\}.
We search for a training epoch in the range \{30, 40, 50\}.
We suggest our total hyperparameter settings in~\Cref{tab:hyperparameter}.
We train our model with a single NVIDIA Titan RTX (24GB) GPU for fine-tuning, and the training time took less than 2 hours.
\vspace{-0.25cm}

\subsection{Experimental Results}
\label{sec:experimental_results}
\Cref{tab:conner_results} reports the results of~\ours~approach.
To show the effectiveness of~\ours~approach, we evaluate our model in the named entity recognition setting in four biomedical domains and compare its performance with that of other methods in different contexts.
Compared to previous biomedical NER models, our approach achieves the best performance compared to all other baselines on three datasets: NCBI-disease~\citep{dougan2014ncbi}, CDR~\citep{li2016biocreative}, and AnatEM~\citep{pyysalo2014anatomical}.
This demonstrates that improving the label agreement of modifiers is effective in a document context.
In addition, the performance gap between BioLM and~\ours~on AnatEM (74.9 vs. 83.5) and Gellus (57.9 vs. 63.4) shows that the label refinement based on the dataset bias, which has low label consistency, is effective.
Although it still lags behind the BiLSTM-CRF model, the gap is reduced.
We provide additional ablation studies of~\ours~approach in the following sections.
\vspace{-0.5cm}

\section{Analysis}
\label{res:ablation}
\subsection{Ablation studies}

\Cref{tab:ablation} shows our ablation result on four biomedical NER benchmarks. 
We evaluate our approach~\ours~by removing its components: 1) distillation (—$L_{distill}$) and 2) the label-refinement process (—$L_{label}$).
The experiments show that~\ours~is effective for all four benchmarks.
Specifically, we observe that the AnatEM and Gellus datasets show significant improvement, demonstrating that our approach~\ours~is effective for datasets with low label consistency on the entities shown in~\Cref{Fig:dataset_bias}.
We also observe that adding each component consistently improves the recall metrics.
These observations correspond to an advantage of~\ours~approach, whereby it decides which token should be refined by relying on the uncertainty threshold $\Gamma$.

\begin{table*}[t]
\processtable{A sample prediction of~\ours~on the AnatEM and CDR datasets. \hl{yellow highlight} refers to the correct prediction and {\color{red}\hl{red}} signifies to the wrong prediction.
\label{tab:qualitative_analysis}} {\resizebox{1.0\textwidth}{!}{{
\begin{tabular}{lllll}
\toprule
\multicolumn{5}{l}{ConNER} \\ \midrule
\multicolumn{5}{l}{\begin{tabular}[c]{@{}l@{}}
\textbf{Dataset}: AnatEM~\citep{pyysalo2014anatomical} \\
\textbf{PMID}: 10420526 \\
\textbf{Title}: {[}Histopathologic examination of \hl{rectal carcinoma}$^{(1)}${]} \\
\textbf{Abstract}: In patients with \hl{rectal carcinoma}, the histopathological evaluation of the \hl{surgical specimen}$^{(2)}$ provides pivotal \\ prognostic and therapeutic information. Important parameters are \hl{tumor}{\color{red}\hl{ site,}} depth of invasion,  histological type and \\ grade, pattern of invasion (diffusely infiltrating versus expanding {\color{red}\hl{margin}}$^{(3)}$), degree of {\color{red}\hl{peritumoral}}\hl{ lymphocytic} infiltration, \\ and \hl{tumor} involvement of \hl{surgical}{\color{red}\hl{ margins}} and {\color{red}\hl{lymph}}\hl{ nodes}$^{(3)}$. Evaluation of the \hl{circumferential}{\color{red}\hl{ (deep, lateral) margin}} \\ is of utmost importance. It should be labeled with ink in the \hl{gross specimen} and should be examined histologically \\ using several \hl{tissue} blocks. The number of {\color{red}\hl{lymph}}\hl{ node metastases} and the total number of lymph nodes examined should \\ be reported. A histological evaluation of the {\color{red}\hl{distal mesorectum}} in its entirety is recommended to detect discontinuous \\ {\color{red}\hl{distal mesorectal}}\hl{ tumor} spread. The histopathological findings should be summarized using the TNM-classification. \end{tabular}} \\ \midrule \midrule
\multicolumn{5}{l}{\begin{tabular}[c]{@{}l@{}}
\textbf{Dataset}: CDR~\citep{li2016biocreative} \\
\textbf{PMID}: 9158667 \\
\textbf{Title}: \hl{Thrombotic}{\color{red}\hl{ complications}} in \hl{acute promyelocytic leukemia} during \hl{all-trans-retinoic acid} therapy \\
\textbf{Abstract}: A case of \hl{acute renal failure}, due to {\color{red}\hl{occlusion of renal vessels}} in a patient with \hl{acute promyelocytic leukemia} \\ (\hl{APL}) treated with \hl{all-trans-retinoic acid} (\hl{ATRA}) and \hl{tranexamic acid} has been described recently. We report a case of \hl{acute }\\ \hl{renal failure} in an \hl{APL} patient treated with \hl{ATRA} alone. This case further supports the concern about \hl{thromboembolic} \\ complications associated with \hl{ATRA} therapy in \hl{APL} patients. The patients, a 43-year-old man, presented all the signs and \\ symptoms of \hl{APL} and was included in a treatment protocol with \hl{ATRA}. After 10 days of treatment, he developed \hl{acute renal }\\ \hl{failure} that was completely reversible after complete remission of \hl{APL} was achieved and therapy discontinued. We conclude \\ that \hl{ATRA} is a valid therapeutic choice for patients with \hl{APL}, although the procoagulant tendency is not completely corrected. \\ \hl{Thrombotic} events, however, could be avoided by using low-dose \hl{heparin}.
\end{tabular}} \\ 
\bottomrule
\end{tabular}}}}{ 
}
\vspace{-0.5cm}
\end{table*}

\begin{figure}[t]
  \centering
  \includegraphics[width=0.235\textwidth]{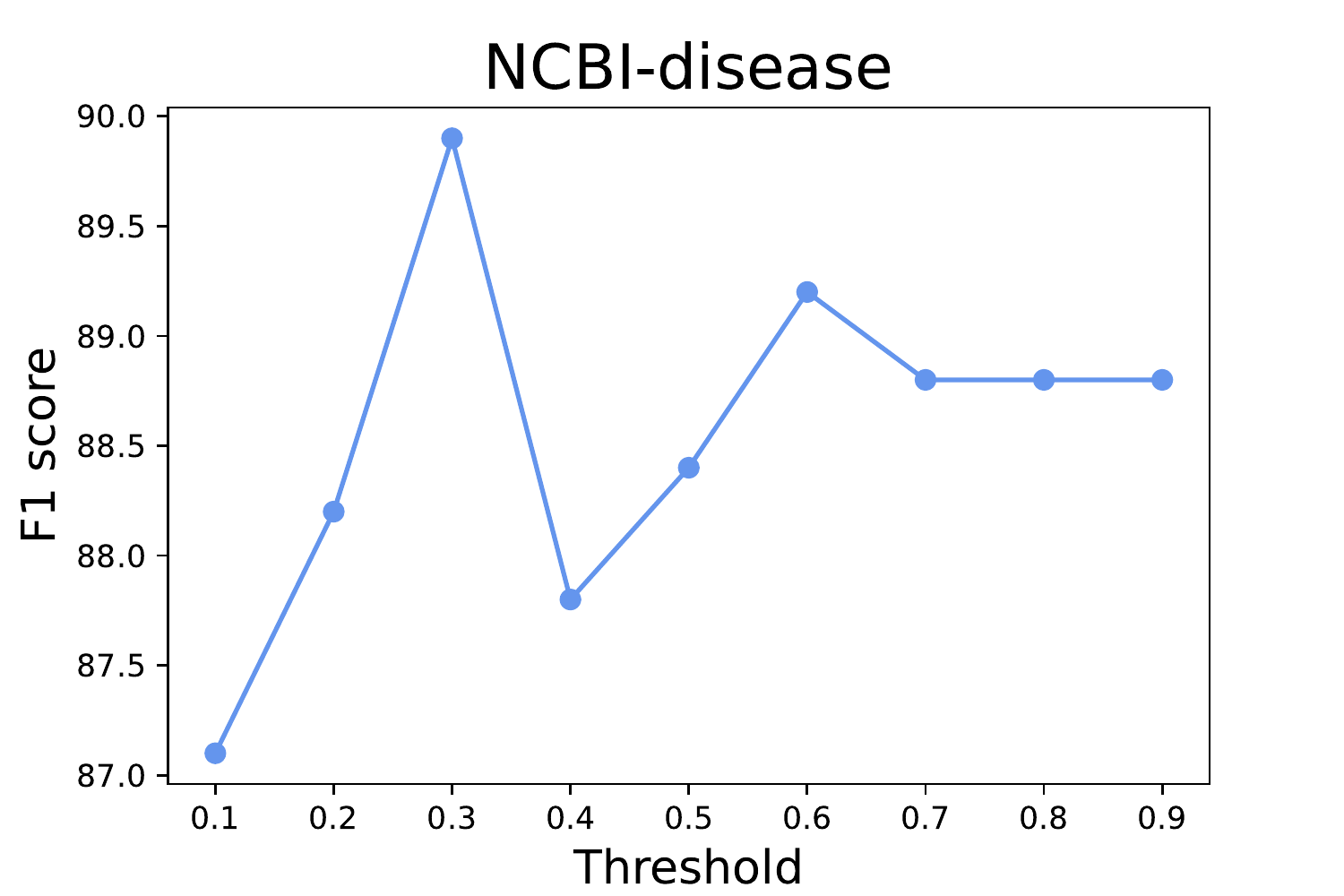} \includegraphics[width=0.235\textwidth]{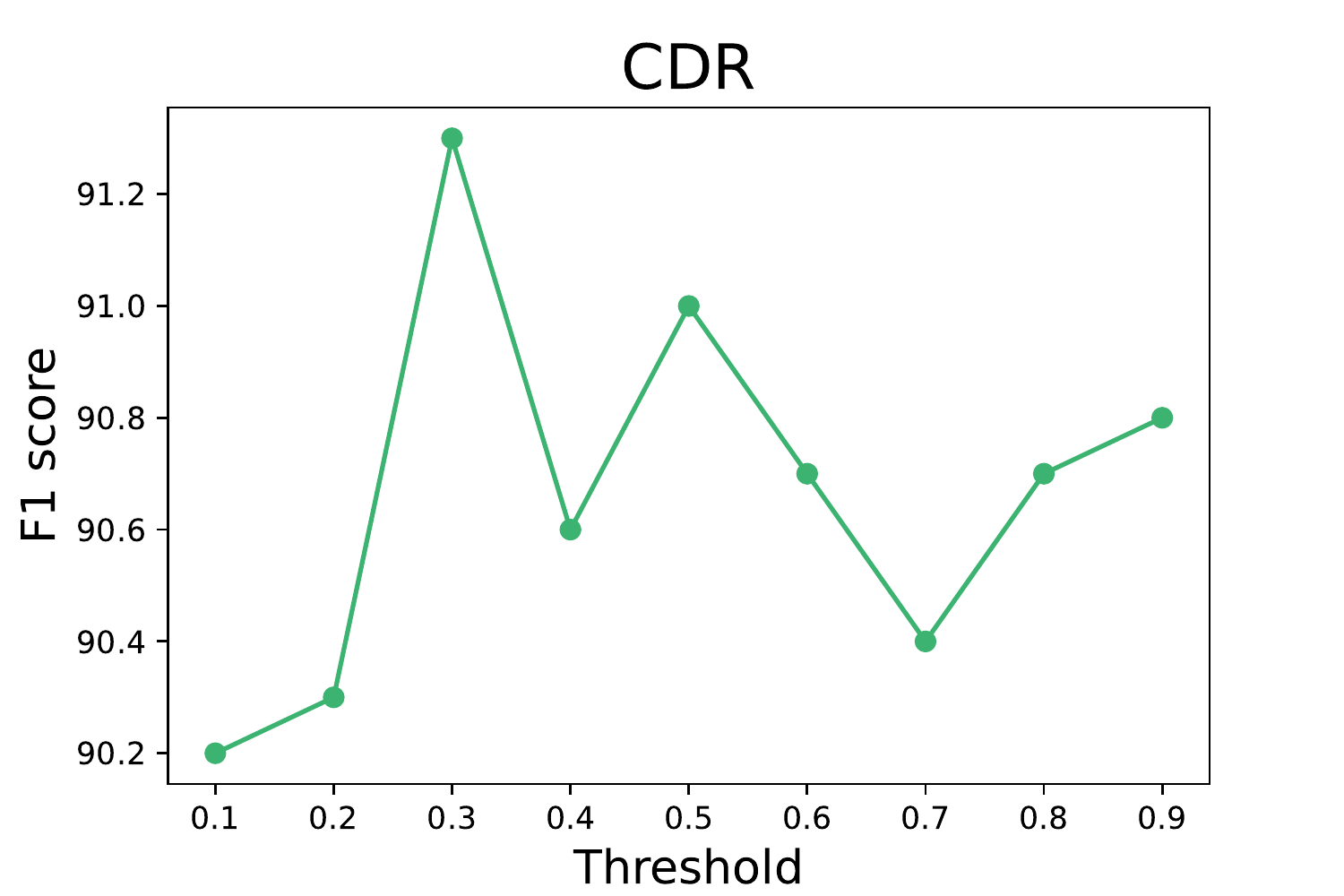} \\
  \includegraphics[width=0.235\textwidth]{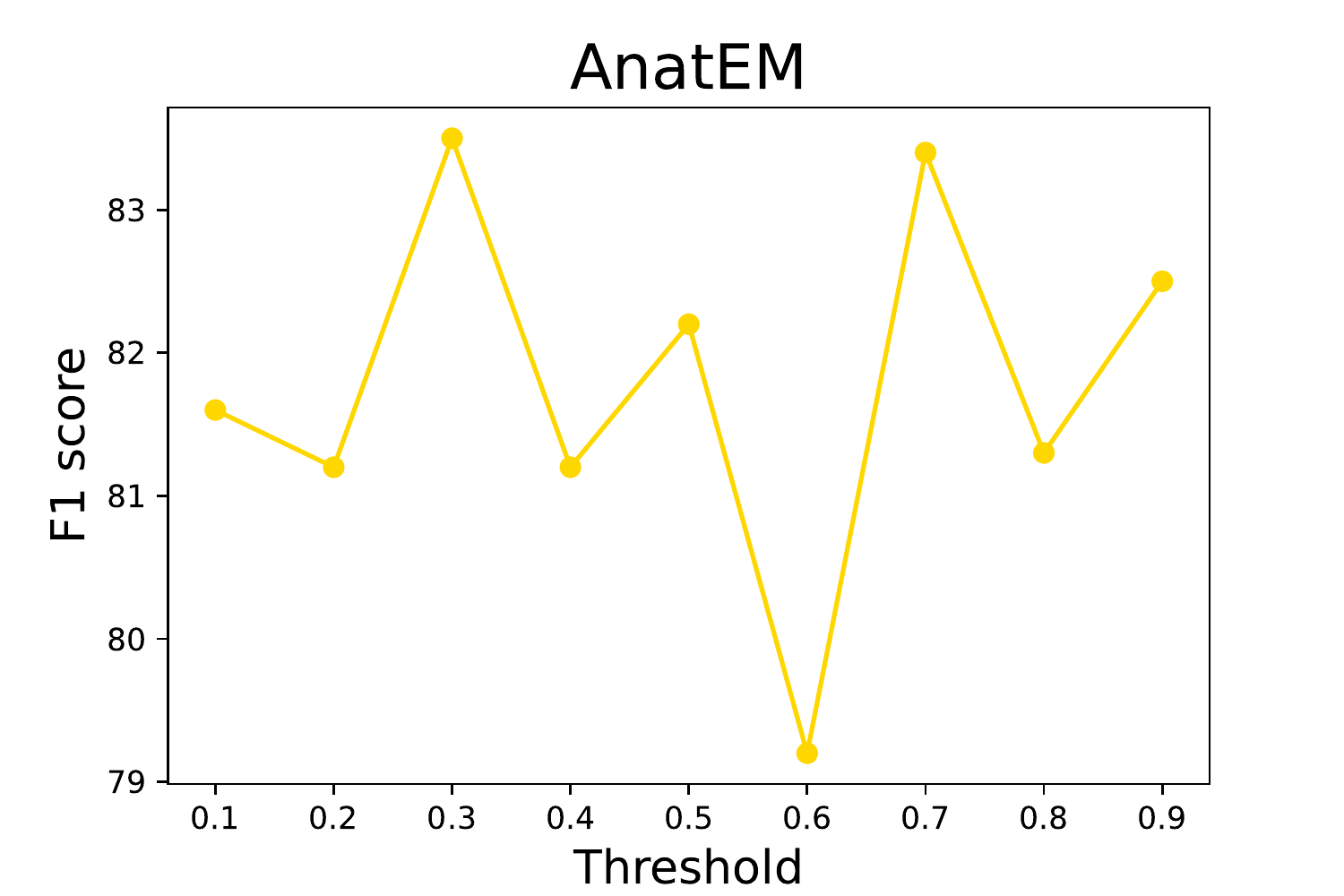} \includegraphics[width=0.235\textwidth]{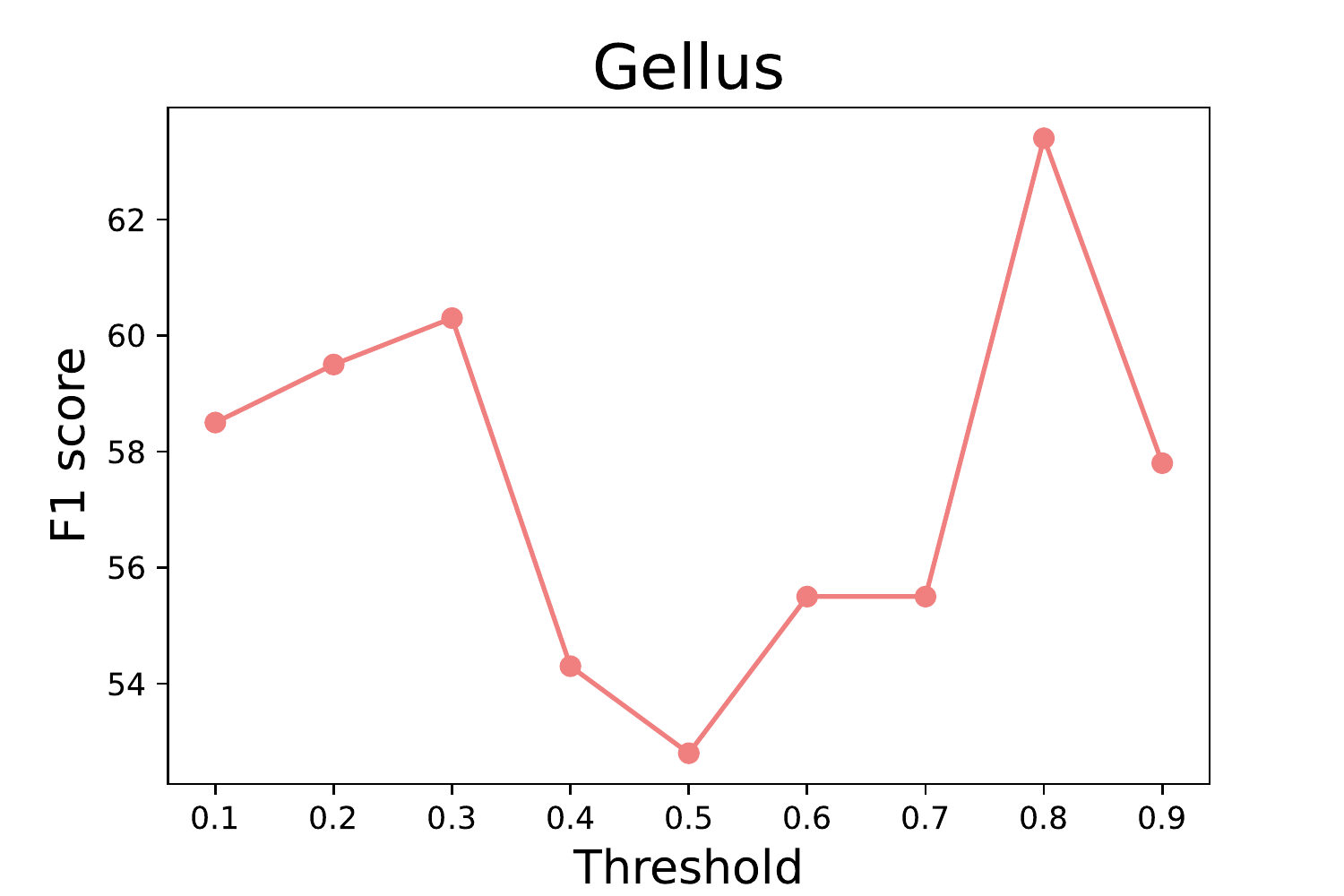} 
  \caption{
  An analysis of uncertainty threshold $\Gamma$. The x-axis refers to the threshold $\Gamma$, and the y-axis denotes the F1 score. The overall results show that $\Gamma=0.3$ results in stable performance on the four biomedical benchmarks.
  }\label{Fig:threshold}
  \vspace{-0.5cm}
\end{figure}

We also provide an interpretation of the threshold $\Gamma$ (see~\Cref{Fig:threshold}).
We observe that~$\Gamma=0.3$ works well in our four benchmarks, except for the Gellus dataset.
When $\Gamma$ is higher than 0.6 in the NCBI-disease dataset, the F1 score yields the same results.
We analyze the situation in which a biomedical pre-trained language model has already demonstrated stable performance in predicting biomedical entities.
We find that~\ours~approach does not have to interfere with the predictions of the main classification layer.
In contrast, in the Gellus dataset,~$\Gamma=0.8$ shows the best performance with high precision and recall metrics.
The results were found to be steady when we compute five times using different seeds.
However, we suggest using~$\Gamma=0.3$ to achieve a stable performance on all benchmarks.

\begin{table}[t]
\processtable{Case studies of inconsistent predictions in document NER models. The below case was taken from the NCBI-disease dataset~\cite{dougan2014ncbi}. $\phi_{tCon}$ refers to the token label consistency. The predictions of BioLM~\citep{lewis-etal-2020-pretrained} and our~\ours~approach are compared. Our~\ours~approach achieves consistent predictions on the modifier examples.
\label{tab:analysis}}{\resizebox{1.0\columnwidth}{!}{{
\begin{tabular}{l c c c c c c c }
\toprule
\multirow{2}{*}{Data} & \multirow{2}{*}{Criteria} & \multicolumn{6}{c}{Modifier Token Examples}\\ \cmidrule{3-8} 
 & & \multicolumn{1}{c}{primary} & \multicolumn{1}{c}{genetic} & \multicolumn{1}{c}{hereditary} & \multicolumn{1}{c}{inherited} & \multicolumn{1}{c}{congenital} & abnormal \\ \midrule
Train & $\phi_{tCon}$ & 0.11 & 0.79 & 0.13 & 0.58 & 0.27 & 0.94 \\ \midrule
\multirow{3}{*}{Test} & $\phi_{tCon}$ & 0 & 0.81 & 0.08 & 0.55 & 0.25 & 1 \\ \cmidrule{2-8} 
 & BioLM & 0\% & 50\% & 50\% & 66\% & 77\% & 44\% \\ \cmidrule{2-8} 
 & ConNER & 100\% & 75\% & 94\% & 100\% & 100\% & 78\% \\ \bottomrule
\end{tabular}}}}{
}
\vspace{-0.5cm}
\end{table}

We also see if our~\ours~approach can improve the label consistency of entities containing modifier tokens.
\Cref{tab:analysis} shows the cases of inconsistent predictions in document NER models. 
We analyze the modifier tokens used as both entity and non-entity tokens depending on the context.
For example, in the training dataset of NCBI-disease, the `\textit{primary}' token is used as an entity only 11 times out of 100 times (i.e., we denote this 0.11 in~\Cref{tab:analysis}).
However, the corresponding token has not occurred as an entity in the test dataset.
We observe that BioLM trained on document context did not predict consistently on these modifier tokens.
Owing to the improvement of label dependency on the tokens within entities, the~\ours~approach achieves consistent predictions on the modifier tokens. 

\subsection{Qualitative analysis}
\label{sec:qualitative_analysis}
\Cref{tab:qualitative_analysis} shows our~\ours~prediction on the AnatEM dataset. 
We investigate which examples could show the advantages and disadvantages of our approach and find three different examples:
(1) Our model consistently predict `\textbf{rectal carcinoma}' mention as an entity. 
Compared to~\ours~prediction, a model without label refinement and distillation process (i.e.,~\ours~—~\{$L_{distill}$,$L_{label}$\}) predicts `\textbf{rectal}' as an entity token in the first appearance and as a non-entity token in the second appearance.
(2) Another surprising example is the token `\textbf{surgical}'.
In the training dataset, the `\textbf{surgical}' token achieved a consistency score of 92.3\% in the entity dictionary.
In other words, it was mostly used as an entity token, accounting for 92.3\% of the total.
The ConNER approach performs well in predicting the `\textbf{surgical}' token as an entity token.
As the intention was to enhance the label dependency of adjectives within entities, it surely works well as per our observations.

However,~\ours~still has several limitations in that it cannot generalize on out-of-density tokens such as `\textbf{mesorectum}' or `\textbf{margin}' that never occur in the training dataset.
\ours~approach cannot predict these tokens as biomedical entities which is a common situation in real-world scenarios.
On comparing the predictions `\textbf{peritumoral lymphocytic}' and `\textbf{lymph nodes}', we can see that the subtoken `\textbf{lymph}' was predicted inconsistently even if they were composed of the same subtoken.
After analyzing the reason why `\textbf{lymph}' subtoken was mispredicted, we interpret that it has a low degree of label agreement in a given paragraph or even in the entire dataset (0.62 of consistency score).
In our future work, we will perform an investigation on ways to make~\ours~more generalizable in out-of-density tokens and more consistent in inconsistent subtokens.

\section{Conclusion}
In this paper, we present~\ours~approach, which enhances label dependency to construct consistent biomedical NER models in document contexts.
We present a label-refinement process and encouragement of label representation to consistently predict biomedical entities.
The~\ours~approach outperforms existing biomedical NER methods in three biomedical domains while providing a view of connecting dataset attributes with a training framework.
We also achieve a powerful performance on low-resource datasets, showing the possibility of adopting our approach for biomedical NER applications.
In our future work, we will attempt to handle out-of-density tokens that never occur in the training datasets, which is a common situation in real-world scenarios.
We expect that our solution will be more sophisticated by not only using golden paragraphs but also utilizing retrieved contexts based on out-of-density tokens.

\section*{Funding}
\vspace{-0.2cm}

\bibliographystyle{natbib}
\bibliography{main}

\end{document}